\documentclass[letterpaper, 10pt, conference]{ieeeconf}      

\IEEEoverridecommandlockouts                              
\usepackage{times} 
\usepackage{graphicx} 
\usepackage{bm} 
\usepackage{amsmath} 
\usepackage{amssymb} 
\usepackage{color} 
\usepackage[ruled,linesnumbered]{algorithm2e} 
\usepackage[short]{optidef} 
\usepackage{gensymb} 
\usepackage{multirow}
\usepackage[mathscr]{eucal}

\usepackage{xparse,soul} 
\soulregister\cite7 
\soulregister\citep7 
\soulregister\citet7 
\soulregister\ref7 
\soulregister\pageref7 

\usepackage[flushleft]{threeparttable}
\usepackage{makecell}
\usepackage{cite}

\usepackage{fancyhdr}
\fancypagestyle{firstpage}{%
    \chead{This paper has been accepted for publication at the 2024 International Conference on Intelligent Robots and Systems.}
    }

\title{Spatial Spinal Fixation: A Transformative Approach Using a Unique Robot-Assisted Steerable Drilling System  and Flexible Pedicle Screw}
\author{Susheela Sharma$^{*1}$, Yash Kulkarni$^{*1}$, Sarah Go$^{1}$, Jeff Bonyun$^{1}$, Jordan P. Amadio$^{2}$,  Maryam Tilton$^{1}$, Mohsen Khadem$^{3}$,\\ and Farshid Alambeigi$^{1}$ \IEEEmembership{Member, IEEE}
\thanks{*These authors contributed equally to this work.}
\thanks{**This work is supported by the National Institute Of Biomedical Imaging and Bioengineering of the National Institutes of Health under Award Number R21EB030796.}
\thanks{$^{1}$S.~Sharma, Y.~Kulkarni, S.~Go, J.~Bonyun, M.~Tilton, and F.~Alambeigi are with the Walker Department of Mechanical Engineering and the Texas Robotics  at the University of Texas at Austin, Austin, TX, 78712, USA. Email: \{sheela.sharma, kulkarni.yash08, sarah.go, jbonyun\}@utexas.edu,   \{maryam.tilton, farshid.alambeigi\}@austin.utexas.edu}.
\thanks{$^{2}$J.~P.~Amadio is with the Department of Neurosurgery, The University of Texas Dell Medical School, TX, 78712. }
\thanks{$^{3}$M.~Khadem is with the School of Informatics, University of Edinburgh, UK. }}

\usepackage{color} 

\begin{document}
\maketitle
\thispagestyle{firstpage}
\pagestyle{empty}

\begin{abstract}
Spinal fixation procedures are currently limited by the rigidity of the existing instruments and pedicle screws leading to fixation failures and rigid pedicle screw pull out.  Leveraging our recently developed Concentric Tube Steerable Drilling Robot (CT-SDR) in integration with a robotic manipulator, to address the aforementioned issue, here we introduce the transformative concept of Spatial Spinal Fixation (SSF)  using a unique Flexible Pedicle Screw (FPS). The proposed SSF procedure enables planar and out-of-plane placement of the FPS throughout the full volume of the vertebral body. 
In other words, not only does our fixation system provide the option of drilling in-plane and out-of-plane trajectories, it also  enables implanting the   FPS inside linear (represented by an I-shape) and/or non-linear (represented by J-shape) trajectories. 
To thoroughly evaluate the functionality of our proposed robotic system and the SSF procedure, we have performed various experiments by drilling different  I-J and J-J drilling trajectory  pairs into our custom-designed L3 vertebral phantoms and analyzed the accuracy of the procedure using various metrics.  
\end{abstract}

\section{Introduction}

Spinal Fixation (SF) is the primary surgical procedure for stabilizing the affected vertebrae and reducing pain caused by movement. This procedure involves using a drilling system utilizing rigid instruments to create straight paths through the pedicles into the vertebral body. Subsequently, rigid instruments like rods and, specifically, Rigid Pedicle Screws (RPS), are employed to secure and immobilize the affected vertebrae. Despite its effectiveness and common use, a significant drawback of RPS fixation is the high risk of screw loosening and pullout, with rates ranging between 22\% and 50\%, even in patients with normal bone mineral density (BMDs) \cite{weiser2017insufficient}. 
 Additionally, RPS fixation often fails to provide sufficient stabilization in osteoporotic vertebrae  (i.e., BMD less than 80 mg/cm³) \cite{weiser2017insufficient,wittenberg1991importance}.
\begin{figure}[!t] 
    \centering 
    \includegraphics[width=1\linewidth]{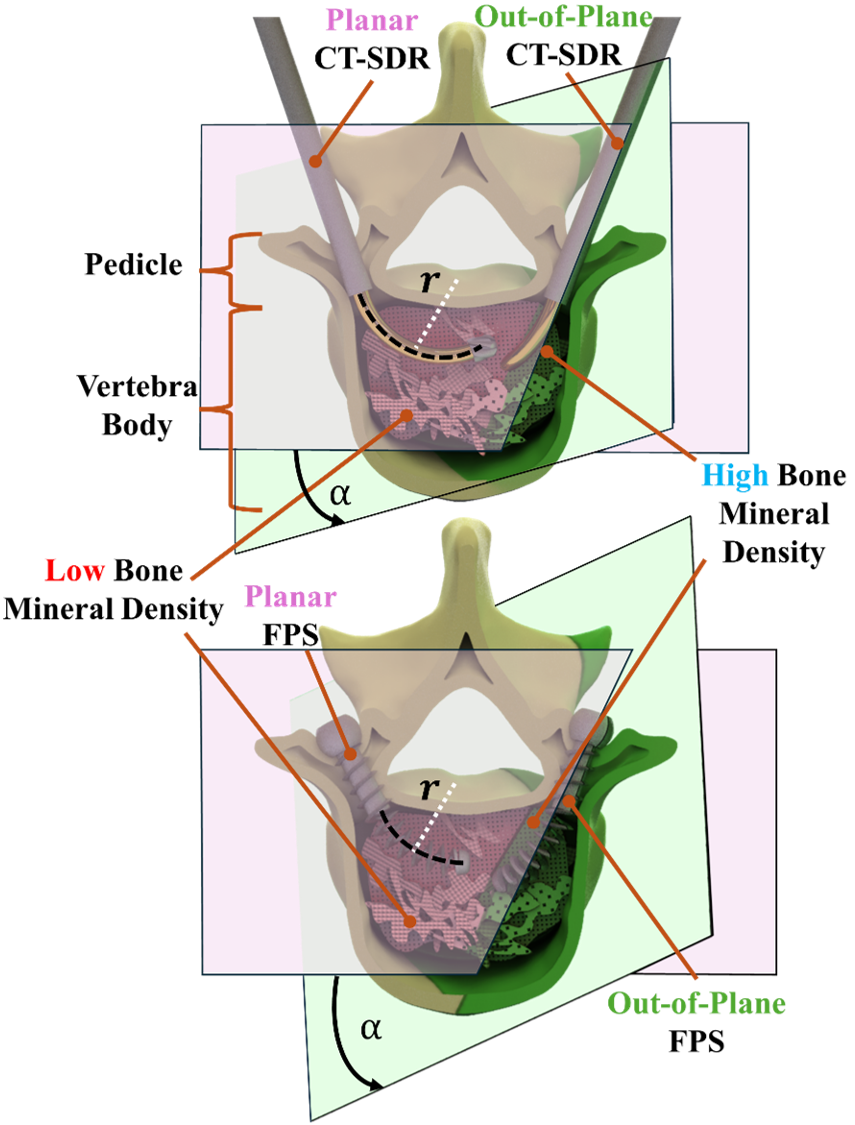}
    \caption{Conceptual illustration of the proposed SSF approach. Top image illustrates the CT-SDR drilling to reach areas of high BMD within the vertebral body in a J-J trajectory configuration. The bottom image illustrates the FPS being fixated within the aforementioned drilled configuration.}
    \label{fig:concept}
\end{figure}
Besides the complex anatomy of vertebrae, the aforementioned issue can be directly attributed to the use of straight rigid instruments  constraining  placement of RPSs solely into a narrow canal located in the midline plane of the vertebral body (see Fig. \ref{fig:concept}) that may limit implantation into  osteoporotic regions. In other words, the existing instruments (both drilling and fixation) cannot not provide the option of \textit{out-of-plane SF (or Spatial SF (SSF))}  and/or avoiding osteoporotic regions within the vertebral body.

In light of the aforementioned issues, several attempts have been made towards improving the success of RPS fixations. Firstly, various methods have been explored by  changing the RPS design parameters such as thread diameter and type \cite{Talu2000PedicleSS} as well as thread pitch \cite{Mehta2012BiomechanicalAO}. Computer-assisted methods have also been explored for pre-operative planning in order to ensure proper screw fixation \cite{Knez2016ComputerAssistedSS}. Nevertheless, these methods still suffer from the  rigidity of screw and lack sufficient access to the vertebral body.

To address the rigidity of the existing drilling instruments and to open more access within the vertebral body, researchers have recently developed three different types of steerable drilling robots (SDRs) for various orthopedics and neurosurgical applications.  Firstly, articulated hinged drilling tools, such as the one introduced by Wang et al. \cite{Hinged2}, opens up a surgeons' access within the vertebra. However, the proposed system is incapable of generating smooth trajectories through the bone for spinal fixation using a potential Flexible Pedicle Screw (FPS). Utilizing tendon-driven continuum manipulators, a tendon-driven SDR has been introduced by Alambeigi et al. \cite{alambeigi2020steerable,alambeigi2017curved}. Despite creating relatively smooth curved trajectories adequate for implant placement, the inherent lack of structural strength in tendon-driven systems poses significant challenges for SF procedures. Recently, we have introduced and successfully evaluated a Concentric Tube Steerable Drilling Robot (CT-SDR), which balances a stronger structural design with the compliance of a curved steering guide for SF procedures using an FPS \cite{Sharma2023TBME, Sharma2023ICRA, Sharma2023ISMR}.

Leveraging our CT-SDR in integration with a robotic manipulator, in this paper and as our main \textit{contributions}, we introduce the transformative concept of SSF using a unique FPS. As conceptually shown in Fig. \ref{fig:concept}, the SSF procedure enables planar and out-of-plane placement of our novel FPS throughout the full volume of the vertebral body and within its high BMD regions. 
In other words, not only our fixation system provides the option of drilling in-plane and out-of-plane trajectories, it also  enables implanting an   RPS and/or FPS inside linear (represented by an I-shape) and/or non-linear (represented by J-shape) trajectories. For example, we can have an, I-I, I-J, or J-J drilling and implantation trajectory configurations using our proposed SSF procedure. To thoroughly evaluate the functionality of our robotic system and the SSF procedure, we have performed various experiments in custom-designed vertebral phantoms and analyzed the obtained outcomes.

\section{Semi-Autonomous Steerable Drilling and Flexible Implant System for SSF Procedure}
\begin{figure}[t]
  \centering
     \includegraphics[width=1.0\linewidth]{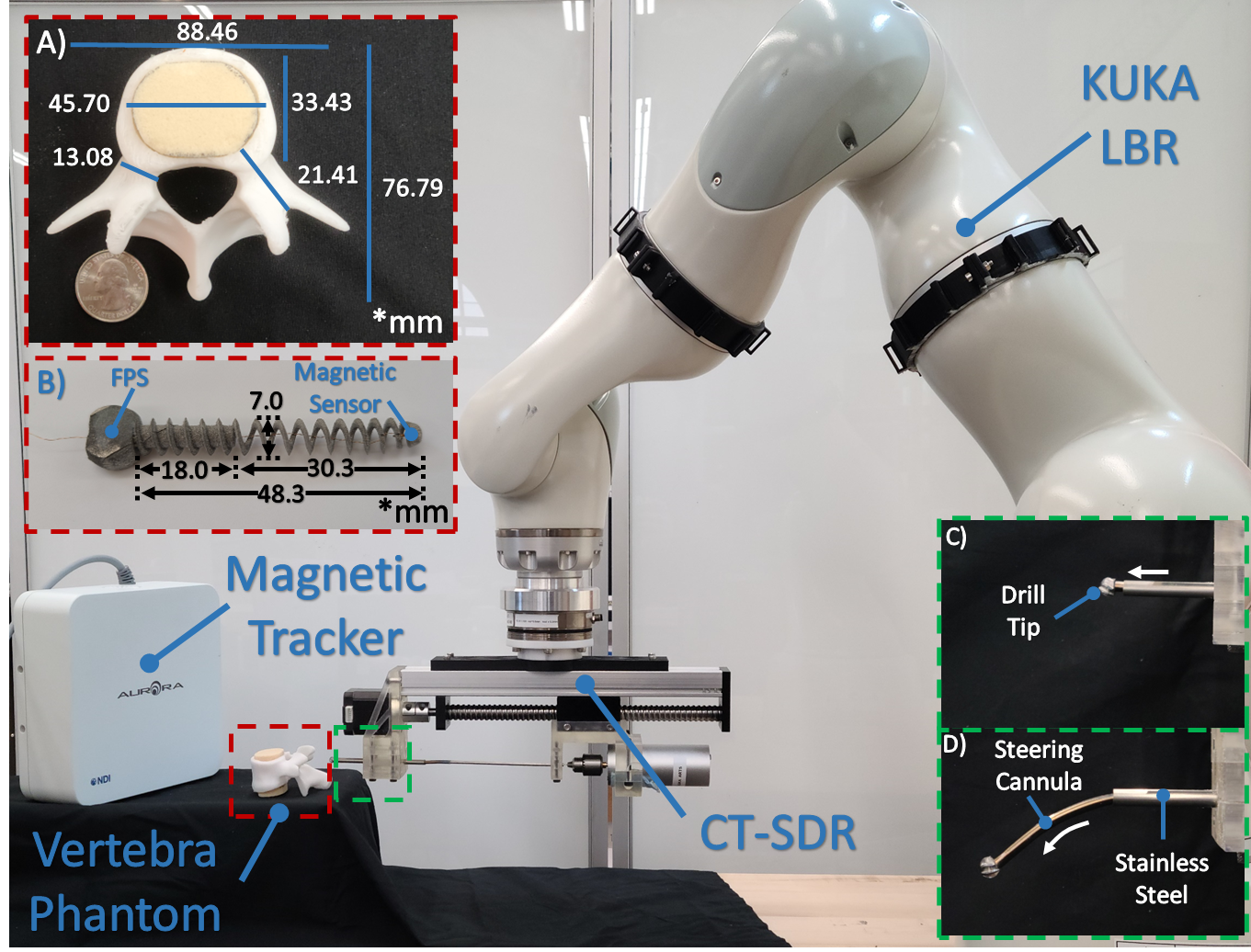}
      \caption{Experimental set-up with CT-SDR, KUKA Robotic Manipulator, and Aurora Magnetic Tracking System. (A) A Close up view of the vertebra phantom with sawbone insert. (B) View of the FPS with magnetic tracking marker placed within the cannulated region. (C\&D) Motion of the CT-SDR's drilling tip when actuated.}
      \label{fig:set-up}
\end{figure}

To enable the proposed SSF procedure and address the aforementioned limitations of rigid tools, we leverage and expand our previously introduced \textit{Biomechanics-Aware Robot-Assisted Steerable Drilling} Framework  in \cite{Sharma2024TBME}. In our proposed SSF approach, first, the \textit{Biomechanics-Aware Trajectory Selection Module} of this framework is used through which a patient's unique Quantitative CT scan is segmented and analyzed to quantitatively ascertain areas of high BMD within the vertebral body. Next, a surgeon  selects the most optimal location for SF and feeds the patient-specific desired drilling/implantation trajectories to the \textit{Semi-Autonomous Robotic Drilling Module} proposed in \cite{Sharma2024TBME}. The passed information includes the I- or J-shape trajectory of the drilling tunnel together with the radius of curvature, $r$, and  orientation angle, $\alpha$, defining the plane of drilling trajectory into the vertebra body, as defined in Fig. \ref{fig:concept}. As illustrated, $\alpha$=0 represents a planar trajectory and  other values represent an out-of-plane drilling/implantation trajectory.
Next, a \textit{Semi-Autonomous Robotic Drilling system} (shown in Fig. \ref{fig:set-up}) is used to   reliably create the aforementioned optimal I- or J-shape trajectories through the patient's vertebra. 
As the final step, we then use our proposed novel FPS and place it into the drilled trajectories. Of note, the FPS is designed to work synergistically with the CT-SDR while safely and reliably conforming to the  created I- and J-shape trajectories. The introduction of an FPS is vital towards enhancing spinal fixation using the proposed SSF technique. The following will define the major components of the proposed SSF procedure in detail and their individual contributions. 

\subsection{Concentric Tube Steerable Drilling Robot}
The CT-SDR, shown in Fig. \ref{fig:set-up}, is comprised of a pre-curved steering cannula, a flexible cutting tool, and a powered actuation unit. The steering cannula are formed from nitinol (NiTi) tubes, heat treated according to the process outlined in \cite{hodgson2001fabrication} to match a pre-determined shape. This allows the surgeon to customize the fixation to the individual patient as defined by the \textit{Biomechanics-Aware Trajectory Selection Module}. The diameter of the used nitinol tubes is 3.61 mm with a wall thickness of 0.25 mm.
The flexible cutting tool is held in the working channel through the center of the CT-SDR and guided by the steering cannula. The flexible tools are created to be extremely compliant and easily guided by the steering cannula present in the system, allowing them to accurately follow the prescribed path.
For the experiments performed in this paper, the flexible cutting tool includes a 6 mm diameter ball nose end mill (42955A26, McMaster-Carr), laser welded to a flexible torque coil 80 mm long (Asahi Intec. USA, Inc.). 
The actuation unit for the CT-SDR utilized in this paper provided a single Degree of Freedom (DoF), allowing the CT-SDR to insert and retract the steering cannula while drilling, a drill motor to control the rotational speed of the cutting tip, and a 3D printed coupling (PLA, Raise3D) to connect the CT-SDR to a robotic manipulator.
\subsection{Semi-Autonomous Robotic Framework Control }
A robotic manipulator provides additional DoFs to place the CT-SDR  and the overall system in the workspace in order to move freely throughout the surgical workspace. For the experiments in this paper, a KUKA LBR med 14 (KUKA, Germany) was utilized for this role as shown in Fig. \ref{fig:set-up}. The KUKA LBR med is a 7 DoF manipulator with load sensing capabilities to allow for a semi-autonomous control strategy. Including a surgeon in the loop of the system adds a layer of safety to the overall procedure as the surgeon can align the manipulator with the patient's vertebra as they see fit and oversee the drilling procedure.
While moving through the procedure, the manipulator was moved through several stages in which different control strategies were implemented. First, an \textit{Admittance Control} strategy was used in which the force applied by the surgeon on the manipulator's end effector guided the robot through the surgical work space through the admittance control technique \cite{li2007spatial}: $\vec{v} = z \mathbf{K} \vec{F}$, where  $\vec{v}\in \mathbb{R}^{6}$ represents the resulting twist vector of the end effector which is formed through a scaling factor, $z\in \mathbb{R}$, the vector of forces and torques from the surgeon's hand on the robot, $\vec{F}\in \mathbb{R}^{6}$, and a diagonal admittance matrix $\mathbf{K}\in \mathbb{R}^{6\times6}$.  The scaling factor was set as $z=15$ (mm/s)/N and the input force $\vec{F}_{initial}$ was pre-processed to introduce a dead zone to reduce any noise felt by the sensors. The diagonal admittance matrix was defined as follows: $\mathbf{K} = diag([1, 1, 1, 0, 0, 0]^\intercal)$. This limited the system's motion to translational so the insertion DoF would follow a specified direction during the drilling procedure. Of note, in the performed procedures, the user is in charge of controlling the drilling plane (i.e.,orientation angle, $\alpha$) based on the out put of the \textit{Biomechanics-Aware Trajectory Selection Module} \cite{Sharma2024TBME}. 
The admittance control stage was followed by an \textit{Autonomous Drilling Control} stage. After being properly aligned by the surgeon, the robotic manipulator performed a straight drilling trajectory by advancing directly into the vertebra through the pedicle. This creates the first half of the drilled trajectory. Upon completion of the straight path, the robot entered a \textit{Stationary Control} stage in which the system held the CT-SDR steady while the second half of the determined curved trajectory is created solely by actuating the CT-SDR.

\subsection{Design and Fabrication of Flexible Pedicle Screws}
In order to ensure the FPS is capable of bending to morph within the curved trajectories created by the CT-SDR, the FPS design must deviate from the typical RPS design. Hence, the FPS design must consider parameters that are critical for the insertion and fixation of the FPS in the vertebral body while also considering parameters that impact the morphability of the implant. The interaction and morphability conditions of the FPS are both critical in dictating the design features of the FPS. Building on work previously done by Alambeigi et al. \cite{Alambeigi2018InroadsTR}, the guiding design criteria for the FPS is introduced below:

\begin{figure}[!t] 
    \centering 
    \includegraphics[width=1\linewidth]{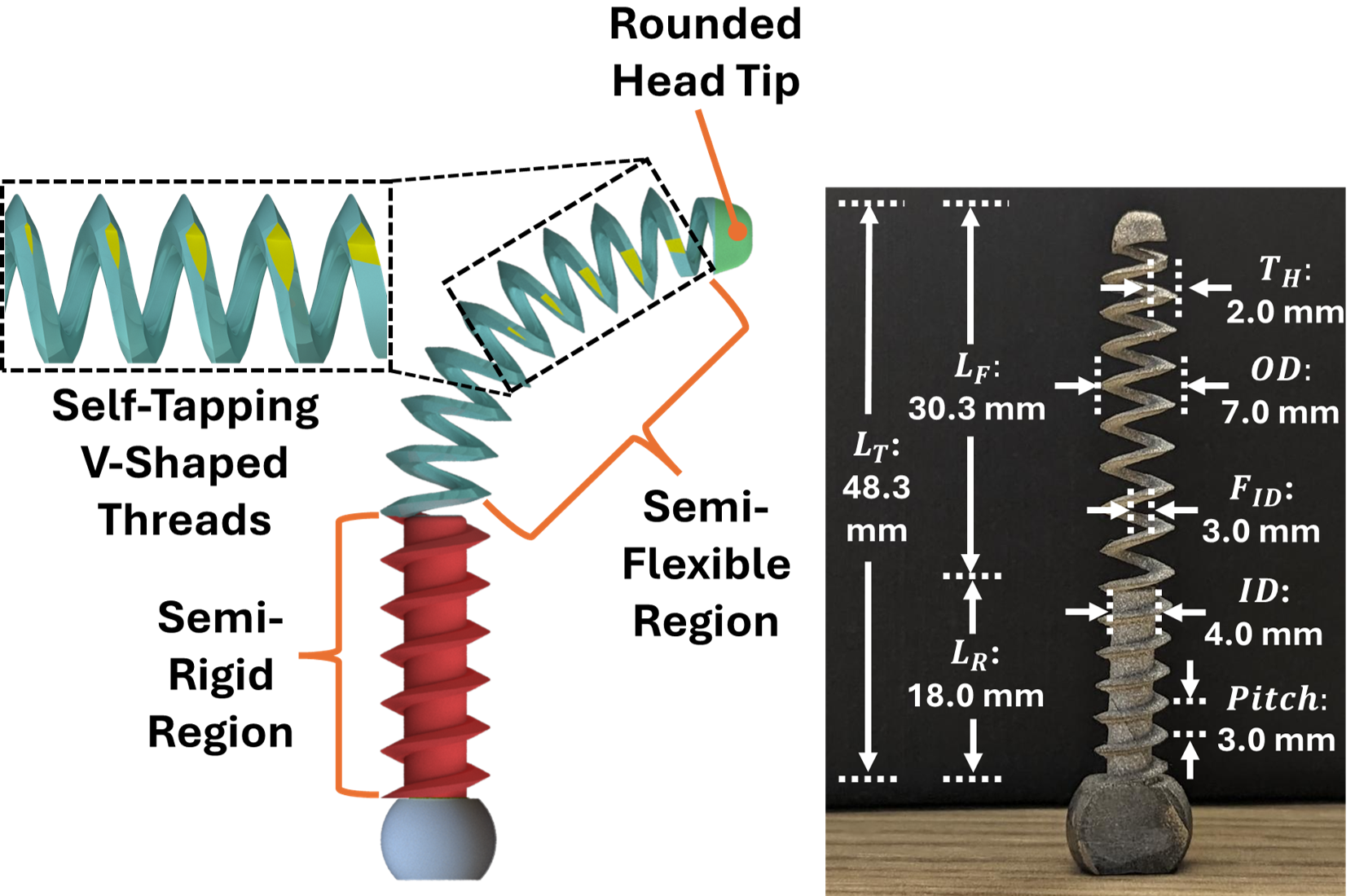}
    \caption{The conceptual design and a fabricated FPS with major design features and parameters labeled on them. The figure on the left illustrates the FPS while it is flexing to matched a proposed J-shape curved trajectory with major design features labeled. The figure on the right illustrates a fabricated FPS in the straight configuration with major design parameters labeled.}
    \label{fig:FPS_AM}
\end{figure}

\textbf{Insertion and Fixation Parameters}: Shown in Fig. \ref{fig:FPS_AM}, the FPS is designed with \textit{Self-Tapping V-Shaped Threads}. These V-shaped threads play a vital role in interacting with the vertebral environment by ensuring easy insertion of the FPS during surgery. Furthermore, the self-tapping threads play a critical role in carving into the bone and tapping it for better engagement of screw threads  \cite{Shea2014DesignsAT}. In order to ensure enhanced fixation, the FPS was designed with five self-tapping V-shaped threads, and a constant 3 mm pitch to ensure sufficient bone-thread fixation occurred. 
The \textit{Rounded Head Tip}  also plays a critical role in ensuring safe interaction between the FPS and the vertebral body. 
The FPS was also designed with an 0.9 mm \textit{cannulated region} going through the middle of it. The cannulated region was created for, if any, bone cement augmentation was required. 

\textbf{Morphability Parameters}: As shown in Fig. \ref{fig:FPS_AM}, we propose a \textit{Semi-Flexible Semi-Rigid} structure for the FPS to meet the I- or J-shape trajectories  created by the CT-SDR. The rigid-shaft plays a critical role in reducing any instability in the pedicle region of the vertebra while the flexible portion of the FPS ensures the curves created by the CT-SDR are precisely matched. These two components ensure the FPS is still flexible enough to reach high areas of BMD in the vertebral body while remaining and providing structural stability to the entire spinal body. Driven by the geometry and the shape of an L3 vertebra, we designed the rigid portion of the FPS to be 18.0 mm long ($L_{R}$) to match the 17.0 mm pedicle length. Furthermore, the flexible portion of the FPS was designed to be 30.3 mm ($L_{F}$) to fit within the 33.4 mm of remaining room in the vertebral body.  Furthermore, the FPS was designed with a 7 mm the outer diameter ($OD$) as it is a common pedicle screw size \cite{Bernard1992PedicleDD}. The FPS was also designed with a 2 mm thread height ($T_H$), a 4 mm inner diameter ($ID$), and a 3 mm inner diameter within the flexible region of the FPS ($F_{ID}$). These parameters were designed holistically to maximize thread-bone interactions while also maximizing bending capability. Of note, in the rigid region, the $T_H$ is slight smaller at 1.5 mm due to its interaction with the rigid $ID$ of the FPS. 
The $ID$ and $F_{ID}$ were finalized at their respective dimension based on the interconnection between these two variables and $T_H$ as well as the fixed $OD$.
All of the aforementioned values are shown in Fig. \ref{fig:FPS_AM}.

\textbf{Fabrication Procedure}: In-line with ASTM F136 \cite{ASTMF136}, Ti-6Al-4V granulates were used for the fabrication of the FPS using the Direct Metal Laser Sintering (DMLS) additive manufacturing process. These granulates were used with a Renishaw AM 250 Laser Melting System (Leuven, Belgium) with a layer thickness of 30 micrometers. Minor post-processing (e.g., sandblasting) was done on the additively manufactured screws. Figure \ref{fig:FPS_AM} shows the fabricated FPS.

\section{Evaluation Experiments and Results}

\subsection{Experimental Set-Up}
Figure \ref{fig:set-up} shows the experimental setup used to perform the SSF procedures. As shown, the setup includes a seven-DoF robotic manipulator (Kuka LBR Medd 14, Germany), the CT-SDR with appropriate flexible drilling instruments, the fabricated FPS sensorized with a 5 DoF magnetic sensor (0.4 mm $\times$ 6 mm) tracked by  a magnetic tracking system (NDI Aurora, Northern Digital Inc.), and a custom-designed L3 vertebral phantom.   To mimic a realistic SF surgical procedure, without loss of generality, for this study, we assumed the \textit{Biomechanics-Aware Trajectory Selection Module}  \cite{Sharma2024TBME} produced J-shape trajectories with   17 mm straight segment and  a radius of curvature equal to 50 mm. To produce this trajectory, the utilized CT-SDR steering guides were heat treated to a 50 mm radius prior to assembly. 
The orientation of the steering guide when assembled into the CT-SDR controlled the final drilled trajectory orientation. 
The custom-designed patient vertebral phantom shown in Fig. \ref{fig:set-up} was designed to match an L3 vertebra. To fabricate the vertebral phantom, the outer cortical layer was fabricated by 3D printing the shell from PLA material (Raise3D), with an 8 mm channel running through the pedicle to act as the alignment channel for the operator. The vertebral body of the phantom was left empty in this shell to allow for the placement of Sawbone (Pacific Research Laboratories, USA) inserts to mimic the vertebral density. We chose PCF 10 for the inserts to mimic medium osteoporotic bone \cite{ccetin2021experimental}. 

\begin{figure*}[t]
  \centering
     \includegraphics[width=1.0\linewidth]{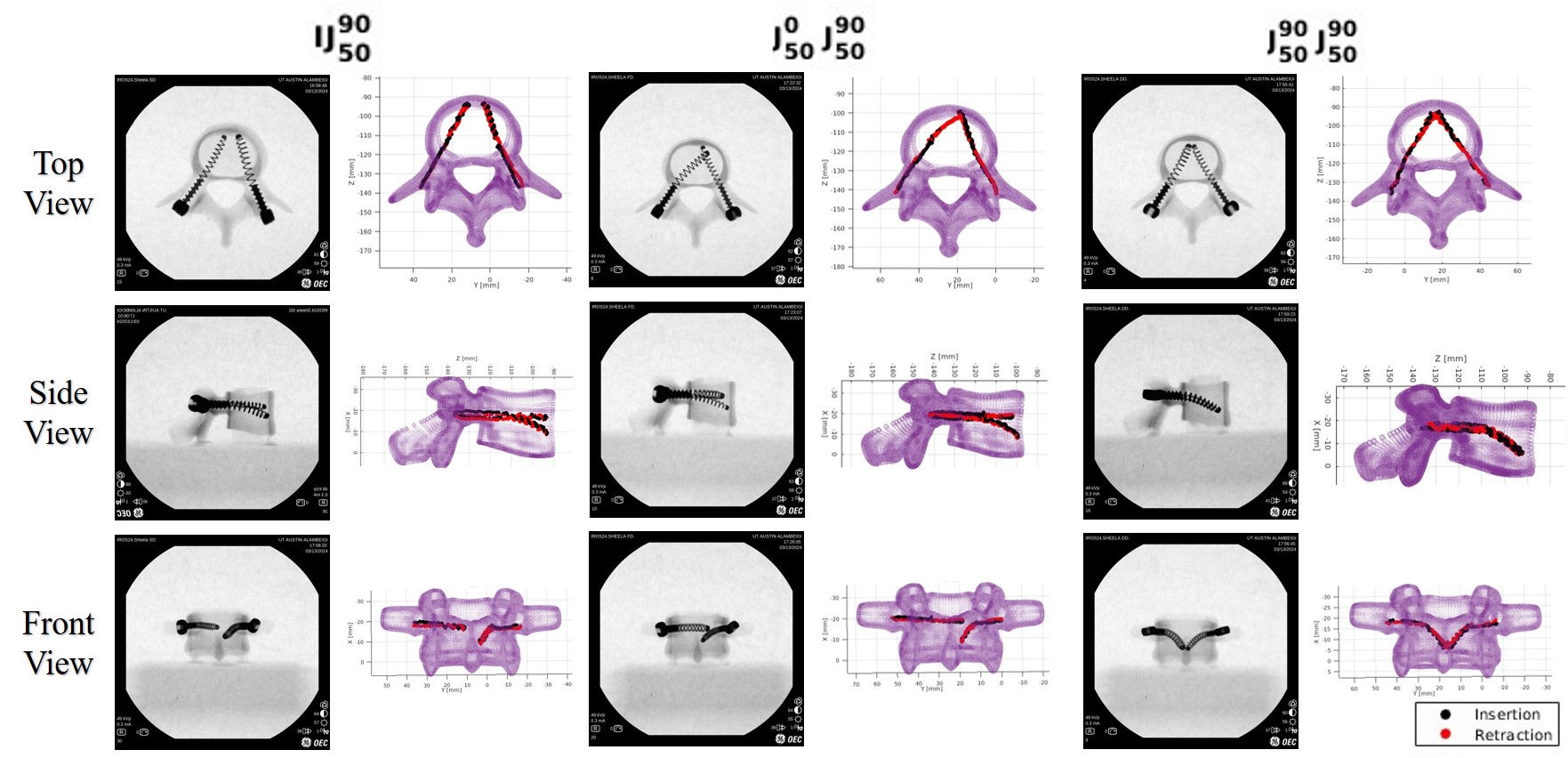}
      \caption{X-ray view of the performed SSF and implanted FPS along paths drilled into the vertebral phantom by the robotic framework. Note: All MATLAB plots/results were created with the 7 mm OD FPS, X-ray images were created with a 7 mm and 6 mm OD FPS.}
      \label{fig:xray}
\end{figure*}

\subsection{Experimental Procedure}
After the vertebral phantom was secured within the surgical workspace, the robotic manipulator began its \textit{Admittance Control} scheme, allowing the surgeon/operator to manually move the tip of the CT-SDR through the workspace. The operator then aligned the CT-SDR with the vertebra phantom based on the desired plane of drilling and the pre-defined orientation angle $\alpha$ (See Fig. \ref{fig:concept}). The CT-SDR's drilling tip was then accelerated to approximately 8250 rpm, before the robotic manipulator began to move to complete the straight segment of the J-shape trajectory using its \textit{Autonomous Control} mode with a translational speed of 1 mm/s. Upon completion of the straight insertion, the manipulator transitioned into its \textit{Stationary Control} scheme as the CT-SDR began the second half of the drilled trajectory. The steering cannula was inserted at a speed of 2 mm/s for 35 mm, creating either a curve (in the case of J-shape trajectories) or a continued straight trajectory (in the case of I-shape trajectories). Upon completion of the drilled trajectory, the CT-SDR's drilling tip was decelerated to approximately 1000 rpm, and the process was reversed to remove the CT-SDR. This process was performed to create the following drilling trajectory pairs: $IJ^{0}_{50}$, $J^{0}_{50}J^{90}_{50}$, and $J^{90}_{50}J^{90}_{50}$ to understand the fixation's capabilities. Of note, in these pairs, $J^{\alpha}_{r}$, represents an out-of-plane drilling trajectory with an angle $\alpha$ and radius of curvature of $r$ for its curved segment (see Fig. \ref{fig:concept}). These orientation angles were selected in order to analyze the extreme cases for placements of the implant within the vertebra for a spatial spinal fixation. After the drilling process, the sensorized FPS was inserted through the drilled tunnels and the insertion trajectory was recorded by the magnetic tracker. 

\subsection{Evaluation Metrics}
After the completion of the drilled trajectories with the robotic framework, the FPS was then evaluated for its ability to follow the created paths and re-creating the desired trajectories determined by the  biomechanics module. The 5 DoF magnetic sensor (Northern Digital Inc.) shown in Fig. \ref{fig:set-up} was placed through the cannulated region of the FPS and fixated within its rounded tip. This allowed for real-time tracking and data collection of the FPS's tip during both insertion and retraction through the drilled tunnels. For better visualization of the drilled trajectories measured by the magnetic tracking system, the Iterative Closest Point (ICP) algorithm \cite{Open3D} was performed on the vertebral phantom, to align the reference frame of the phantom's 3D model with the frame of the magnetic tracking system. This allowed for the creation of the plots shown in Fig. \ref{fig:xray} with both the vertebra and screw trajectory, and to provide the trajectory location where it transitions from straight to curved segment. After the transition point is located, the curved section is analyzed with a best fit to determine the curvature of the placed FPS and compare it with the desired curvature. The insertion and retraction were measured 3 times each for both  $J^{0}_{50}$ and $J^{90}_{50}$ trajectories. For further visualization of the placed implants, the vertebra was then observed with a C-arm X-ray machine (OEC One CFD, GE Healthcare) with two FPS placed into the trajectory combinations. These obtained results  can  be seen in Fig. \ref{fig:xray}. Also, Table \ref{table:1} summarizes the calculated errors between the obtained and desired SSF trajectories. 

\section{ Discussion}
 The safety of the proposed SSF procedure is ultimately determined by the ability of both the CT-SDR and the FPS to follow the desired I- or J-shape trajectories determined by the  \textit{Biomechanics-Aware Trajectory Selection Module} \cite{Sharma2024TBME}. The shown X-ray views for  different drilling pairs in Fig. \ref{fig:xray}, clearly demonstrates the potential of the proposed SSF approach using our proposed FPS. As can be observed, the FPSs have successfully morphed into planar and out-of-plane drilled trajectories.  Also, to quantitatively measure the accuracy of these SSF procedures, we analyzed the trajectories of the placed FPS with a magnetic tracker and compared the results to ideal trajectory. As shown in Fig. \ref{fig:xray} and reported in Table \ref{table:1}, for the two extreme orientations $\alpha$=0 and $\alpha$=90 (i.e., $J^{0}_{50}$ and $J^{90}_{50}$) the behavior of the CT-SDR and implant were very similar.
The average  radius of curvature for all of the measured tests was 49.43$\pm$2.43 mm, corresponding to an average error of 1.14\% when compared to the patient's bio-mechanically defined 50 mm ideal radius. 
The low standard deviation seen across these tests corresponds to the accuracy of the system, independent of the orientation. This leads us to believe the system is capable of placing an FPS within the vertebral body at any orientation between the mid-plane ($\alpha=0$) and pointed downward (i.e., $\alpha=90$) shown in Fig. \ref{fig:xray}.  Across all measured tests, both $J^{0}_{50}$ and $J^{90}_{50}$, the maximum error was seen in a $J^{90}_{50}$ test at 45.00 mm. 
The highest accuracy was seen in the $J^{0}_{50}$ trajectory tests, with a measured radius of 50.03 mm. Overall this shows great accuracy in the ability of the CT-SDR and the FPS to follow the expected trajectory. 

\begin{table}
\centering
\begin{threeparttable}
\caption{Experimental Results across trials completed for each SSF trajectory }
\label{table:1}
\setlength\tabcolsep{0pt} 
\begin{tabular*}{1\columnwidth}{@{\extracolsep{\fill}} l ccccc}
\Xhline{1.25pt}
Trajectory & Ideal & $J^{0}_{50}$ & $J^{90}_{50}$ & Combined\\
\Xhline{1.25pt}
Radius  of \\Curvature & 50.00  & 49.28 $\pm$ 1.70 & 49.61 $\pm$ 3.55 & 49.43 $\pm$ 2.43\\
\Xhline{0.25pt}
{} \\ Error & 0  & 1.44\% & 0.78\% & 1.14\%\\
\Xhline{0.25pt}
 ICP \\ RMSE & 0  & 0.8291  & 0.8221 & x\\

\Xhline{1.25pt}
\end{tabular*}
\smallskip
\scriptsize
\begin{tablenotes}
\item
All units are in mm.
\end{tablenotes}
\end{threeparttable}
\end{table}

Figure \ref{fig:xray} also shows the  trajectories gathered from the combination of the FPS's trajectories through the vertebral phantom and the 3D CAD model of the same phantom after performing an ICP registration. The accuracy of these plots are integral to the visualization of these plots and towards accurately placing the location of transition between the straight and curved portions of the trajectory. The utilized ICP algorithm from \cite{Open3D} provides the RMSE calculated between the resolved points after transformation and the goal points on the original vertebra. For the measured trajectories, the RMSE values for the performed ICP algorithms are listed in Table \ref{table:1}. The relatively low errors in the correspondence set of the collected magnetic tracker points and the 3D CAD model show the accuracy of the used algorithm and the performed SSF procedures. 

Another important factor contributing towards adaptation of the proposed SSF procedure by clinicians is the overall drilling time compared with the existing SF procedure with rigid instruments. This is greatly controlled by the speed at which the drilling system is inserted  into the vertebral body while reliably and accurately cutting through material without experiencing buckling or other failures. The CT-SDR in this paper was advanced via the robotic manipulator for the straight segment of the J-shape trajectory at a speed of 1 mm/s, before advancing through the curved trajectory at a speed of 2 mm/s, leading to a final cutting time of 34.5 s for a completed trajectory. Other steerable drilling devices report significantly higher cutting times, as systems based on tendon-driven designs are extremely compliant and buckle/deform easily in the presence of external forces. Alambeigi et al. \cite{alambeigi2017curved} show cutting times of over 300 s when drilling trajectories of comparable length. Therefore, besides the phenomenal accuracy of the proposed SSF procedure, this unique approach does not significantly change the drilling time and surgical workflow of the current SF procedures. 

\section{Conclusion and Future Work}
In this paper, we  proposed a novel SSF approach that takes advantage of our previously proposed CT-SDR system and a novel FPS to enable fixation into previously inaccessible high BMD regions throughout the vertebral body. 
For the first time to our knowledge, we (i) introduced the transformative concept of SSF, (ii) introduced the design, fabrication, and evaluation of a unique FPS, and (iii) successfully assessed the performance of the proposed SSF procedure by drilling different  I-J and J-J drilling trajectory  pairs into our custom-designed L3 vertebral phantoms.
An average error of 1.14\% was reported across all testing, resulting in an actual radius of curvature of 49.43 mm compared to the ideal goal of 50 mm. These results prove the feasibility of the proposed SSF approach for out-of-plane FPS placement within any location inside the vertebral body. In the future, we will expand the evaluation of the SSF approach to animal bone and cadaveric specimens. We will also plan to perform an image-guided autonomous SSF procedures using the proposed robotic system.

\bibliographystyle{./IEEEtran}
\bibliography{./root}

\end{document}